\documentclass[]{spie}  %>>> use for US letter paper
\usepackage{amsmath,amsfonts,amssymb}
\usepackage{graphicx}
\usepackage{booktabs} 
\usepackage[colorlinks=true, allcolors=blue]{hyperref}
\usepackage{multirow}

\title{Noise Injection: Improving Out-of-Distribution Generalization for Limited Size Datasets}

\author{Duong Mai, Lawrence Hall}
\affil{Bellini College of Artificial Intelligence, Cybersecurity and Computing, University of South Florida, Tampa, FL, 33620 USA}
\authorinfo{Further author information (Send correspondence to Duong Mai) \\ Duong Mai: E-mail: duongmaixa1207@gmail.com, Telephone: 1 656 200 4352\\ Lawrence Hall: E-mail: lohall@usf.edu, , Telephone: 1 813 974 4195}
\begin{document} 
\maketitle

\begin{abstract}
Distribution shifts in machine learning (ML) models occur when the training data distribution differs from the test distribution.  This causes model performance to  become inconsistent in real-world deployment. Unlike natural image datasets which can be crawled from massive internet sources to approximate many common domains, medical imaging data is protected by HIPAA and presents significant challenges for collecting sufficiently large and diverse training datasets to the variety of acquisition and demographic differences that the test distribution may reflect. In practice, deep learning (DL) disease classification models are often trained on limited collections of images from one or two sources and then deployed across both internally and externally where images may exhibit wildly differing physical imaging qualities due to acquisition, transmission, or storage protocols. In such cases, while the underlying pathology characteristics may remain consistent across sites, DL models often overfit by choosing shortcuts, or spurious physical imaging artifacts, over appropriate diagnostic biomarkers to quickly maximize their in-distribution (ID) performance. This shortcut learning, however, leads to unstable performance in out-of-distribution (OOD) settings as such learned features are prone to variations across domains. In this study, we propose noise injection, a simple yet practical noise-based data augmentation pipeline, to partially address this inconsistency challenge. Specifically, for a COVID-19 vs. pneumonia classifier trained on chest radiographs (CXRs), we leverage several common imaging noise artifacts -- \textbf{Gaussian, Speckle, Poisson}, and \textbf{Salt \& Pepper} noise -- to improve model robustness to variations present in external hospital settings not covered in the training set. We evaluate our methods by comparing ID and OOD performance under standard training versus our proposed approach. Empirical results demonstrate substantial improvements in the performance gap between the two settings across standard classification metrics (AUC, F1, accuracy, recall, and specificity), averaged over 30 runs with different random seeds. In addition, they suggest an appropriate training data composition can yield decent OOD generalizability even if the dataset may be limited in size. Our source code to reproduce the experiments is publicly available at \url{https://github.com/Duongmai127/Noisy-ood}
\end{abstract}


\section{INTRODUCTION}
\label{sec:intro}

% PARAGRAPH 1: Broader Context - Distribution Shifts in Medical Imaging
Distribution shift occurs when test data differs systematically from training data, and it often poses fundamental challenges to a trustworthy, robust and ubiquitous clinical deployment of Deep Learned (DL) models in medical imaging. Unlike natural images where large-scale datasets can be crawled across the entire internet, medical imaging faces strict privacy constraints and data sharing barriers that restrict both the scale and diversity of training datasets \cite{gutbrod2025openmibood}. Consequently, it is often impractical to ensure that all potential deployment domains (e.g., different hospitals, scanner manufacturers, or acquisition protocols) are adequately represented during training \cite{koh2021wilds}. Due to this limitation, medical imaging models are particularly subject to overfitting on ``shortcuts", or source-specific artifacts that fail to generalize to new clinical environments \cite{lopez2021current}. Especially for safety-critical applications such as COVID-19 detection from chest radiographs (CXRs), distribution shifts and shortcut learning pose serious risks to both clinical trust and patient safety as reliable performance is more essential than ever in these settings.

There are multiple types of distribution shifts in medical imaging \cite{hong2024out}, including variations in patient demographics, disease prevalence, imaging equipment, and institutional protocols. These shifts are particularly complex in multi-hospital deployments since each institution may employ different scanners, positioning views, and image processing pipelines. In the context of COVID-19 pneumonia detection from CXRs, recent evidence has demonstrated that deep learning models tend to exploit hospital-specific shortcuts such as scanner artifacts, text overlays, patient positioning markers, or image processing signatures rather than learning clinically relevant pathological biomarkers\cite{momeny2021learning, degrave2021ai, ahmed2021discovery}. These spurious correlations arise from differences in acquisition protocols, transmission standards, and storage formats across institutions\cite{ahmed2021discovery}. Consequently, models that achieve high performance on in-distribution (ID) test sets from the same sources as training data often exhibit substantial performance degradation when evaluated on out-of-distribution (OOD) data from new hospitals or imaging systems\cite{hong2024out, koh2021wilds, ahmed2021discovery, lopez2021current}.

The domain generalization literature proposes two primary strategies to address distribution shift: (1) data augmentation techniques that simulate variations expected in test domains, and (2) regularization methods that encourage feature alignment across multiple source domains during training \cite{li2020domain}. For scenarios with limited source diversity i.e. single-source settings common in resource-constrained healthcare environments, data augmentation represents the more practical approach. Among augmentation strategies, noise injection is particularly relevant for CXRs analysis, as imaging noise is ubiquitous and varies substantially across scanners, technicians, and institutional protocols\cite{momeny2021learning, akbiyik2023data}. However, existing evaluations of noise-based augmentation have primarily focused on robustness to covariate shifts within the same source, such as different X-ray machine configurations or patient positioning from hospitals already represented in the training set\cite{gutbrod2025openmibood}. The impact of training-time noise augmentation on generalization to genuinely external sources under the realistic constraint of limited training data size and source diversity remains substantially underexplored. This gap is critical, as the effectiveness of augmentation strategies may depend on the composition and diversity of available training sources.

In this work, we systematically investigate the effect of training-time noise injection on OOD generalization for COVID-19 CXR classification under realistic constraints of limited data size and source diversity. Our primary objective is to reduce the performance gap between ID and OOD evaluations, thereby providing radiologists and clinical decision-makers with more reliable and trustworthy performance expectations across deployment sites. Beyond evaluating augmentation effectiveness, we analyze how the composition and dissimilarity of training sources influences a model's capacity to learn generalizable features versus exploitable shortcuts. Our investigation addresses a practical clinical scenario: institutions seeking to deploy AI models trained on their own limited data or small consortium while maintaining acceptable performance when the model encounters data from new hospitals with different imaging protocols.

We make two primary contributions relevant using COVID-19 detection from limited CXR datasets with restricted source diversity. First, we demonstrate that noise-based data augmentation can substantially reduce the ID-OOD performance gap across key clinical metrics including AUC, accuracy, F1 score, recall, and specificity, without requiring model fine-tuning on target sources. Second, we reveal that training dataset composition plays important roles in model robustness, similar to how a larger dataset does not necessarily correlate with better model performance \cite{shen2024data}. As the dissimilarity among training sources increases, the benefits of training-time noise injection for OOD generalization diminish, which has important implications for data acquisition strategies and suggests that the value of augmentation must be considered in conjunction with the diversity of available training data.

\section{METHODS}
\label{sec:methodology}
To demonstrate the effect of noise injection on model generalization, we train a DL model to classify COVID-19 versus non-COVID-19 pneumonia from CRX images. In healthcare, privacy constraints often constrain the size and diversity of training data. \cite{gutbrod2025openmibood}. Therefore, our study stimulates a common real-world scenario by training our model on a small subset of data from a single medical network and evaluating it on external data of the same pathology. This approach allows us to assess the model's ability to generalize to unseen data sources.

\begin{table}[ht!]
\centering
\caption{Dataset Composition and Splits}
\label{tab:dataset_splits}
\begin{tabular}{|l|l|l|c|c|}
\hline
\rule[-1ex]{0pt}{3.5ex} \multirow{1}{*}{\textbf{Data split}} & \textbf{Data sources} & \textbf{Class} & \textbf{Image Count} & \textbf{Total} \\
\hline
\rule[-1ex]{0pt}{3.5ex} \multirow{2}{*}{Training} & BIMCV & Covid-19 & 245 & \multirow{2}{*}{509} \\
% \cline{2-4}
\rule[-1ex]{0pt}{3.5ex} & Padchest & Pneumonia & 264 & \\
\hline
\rule[-1ex]{0pt}{3.5ex} \multirow{2}{*}{Validation} & BIMCV & Covid-19 & 27 & \multirow{2}{*}{56} \\
% \cline{2-4}
\rule[-1ex]{0pt}{3.5ex} & Padchest & Pneumonia & 29 & \\
\hline
\rule[-1ex]{0pt}{3.5ex} \multirow{2}{*}{ID Test} & BIMCV & Covid-19 & 38 & \multirow{2}{*}{97} \\
% \cline{2-4}
\rule[-1ex]{0pt}{3.5ex} & Padchest & Pneumonia & 59 & \\
\hline
\rule[-1ex]{0pt}{3.5ex} \multirow{4}{*}{OOD Test} & Arkansas & \multirow{2}{*}{Covid-19} & 75 & \multirow{4}{*}{849} \\
% \cline{2-2} \cline{4-4}
\rule[-1ex]{0pt}{3.5ex} & Germany & & 155 & \\
% \cline{2-4}
\rule[-1ex]{0pt}{3.5ex} & NIH & \multirow{2}{*}{Pneumonia} & 205 & \\
% \cline{2-2} \cline{4-4}
\rule[-1ex]{0pt}{3.5ex} & Chexpert & & 414 & \\
\hline
\end{tabular}
\end{table}
\subsection{Data}
\label{sec:data}
Our experimental design divides the data into 2 groups: \textbf{ID} data for training, validation, and testing and \textbf{OOD} data reserved for testing generalization. For simplicity, we refer to the classes as \textit{COVID-19} and \textit{Pneumonia} (for non-COVID-19 cases), with labels 0 and 1, respectively. The data selection, naming conventions, and preprocessing pipelines are adapted from the methodologies presented in Ref.~\citenum{ahmed2021discovery}. 

\subsubsection{Data Selection}
\label{sec:dataselection}
For the \textbf{ID} dataset, we selected CXR images (AP or PA views) from the Valencian Region Medical ImageBank (BIMCV) network (Spain), specifically BIMCV-COVID-19 + \cite{vaya2020bimcv} and Pachest\cite{bustos2020padchest} for COVID-19 and pneumonia, respectively. 

For \textbf{OOD} evaluation, on the other hand, we sourced data from multiple medical institutions. While COVID-19 cases came from COVID-19-AR\cite{desai2020chest} (USA) and V2-COV19-NII\cite{Winther2020} (Germany), Pneumonia cases were composed of NIH\cite{wang2017chestx} (USA) and Chexpert\cite{irvin2019chexpert} (USA). A detailed summary of the data splits is provided in Table~\ref{tab:dataset_splits}.

\subsubsection{Data Preprocessing}
\label{sec:datapreprocessing}
The raw CXR images exhibit widely different variations in rotation, size, and imaging artifacts. To standardize the input and mitigate the undesirable impact of shortcuts, we implemented a two-step preprocessing pipeline. First, we isolated the chest area in each image using HybridGNet \cite{gaggion2022improving}, a pretrained neural net for lung segmentation. Second, the resulting cropped images were then normalized to an 8-bit resolution, duplicated to 3 channels, and resized to 224x224 dimensions. This final step ensures compatibility with the ResNet-50 feature extractor used in our classifier (See Sec.~\ref{sec:training}).

\subsection{Experimental Design}
\label{sec:experiment}
\subsubsection{Noise injection}
\label{sec:noise}
To enhance robustness, we employed a noise-based data augmentation strategy, also known as training-time noise injection. Following the recommendations of Ref.~\citenum{momeny2021learning}, we applied 4 types of noise that stimulate artifacts from acquisition, transmission or storage: \textbf{Gaussian}, \textbf{Speckle}, \textbf{Poisson} and \textbf{Salt \& Pepper}. The parameter values for each noise distribution, as summarized in Table~\ref{tab:noise_params_consistent}, were selected based on their common implementations across several image processing frameworks such as Scikit-Image or TorchVision.

\begin{table}[ht!]
\centering
\caption{Noise Augmentation Parameters}
\label{tab:noise_params_consistent}
\begin{tabular}{|l|l|c|c|}
\hline
\rule[-1ex]{0pt}{3.5ex} \textbf{Type of Noise} & \textbf{Parameters} & \textbf{Range} & \textbf{Value} \\
\hline
\rule[-1ex]{0pt}{3.5ex} \multirow{2}{*}{Gaussian} & Mean & [0,1] & 0.0 \\
% \cline{2-4}
\rule[-1ex]{0pt}{3.5ex} & Variance & [0,1] & 0.01 \\
\hline
\rule[-1ex]{0pt}{3.5ex} \multirow{2}{*}{Salt 'n Pepper} & Density & [0,1] & 0.05 \\
% \cline{2-4}
\rule[-1ex]{0pt}{3.5ex} & Salt-Pepper Ratio & [0,1] & 0.5 \\
\hline
\rule[-1ex]{0pt}{3.5ex} Speckle & Variance & [0,1] & 0.01 \\
\hline
\rule[-1ex]{0pt}{3.5ex} Poisson & N/A & N/A & N/A \\
\hline
\end{tabular}
\end{table}
\subsubsection{Training Details}
\label{sec:training}
Given our limited training data, we utilized transfer learning using a ResNet-50 architecture. We froze the pre-trained feature extractor and fine-tuned only the classification head. Our implementation used the official TorchVision ResNet-50 with \verb|IMAGENET1K_V2| pretrained weights, resulting in a total of 174K trainable parameters in the classification head.

We trained our model using a binary cross-entropy loss function and an Adam optimizer with a learning rate of $10^{-4}$, which was adjusted using an exponential decay scheduler. The maximum number of epochs was 100, and early stopping was done by monitoring the AUC score in the validation set. Once the validation AUC stopped improving for 5 consecutive epochs, training ceased, and the best model checkpoint was saved.

The complete training source code can be found on our GitHub repo \url{https://github.com/Duongmai127/Noisy-ood}

\subsubsection{Experiment Design}
\label{sec:experiment_design}
To assess the impact of noise injection on model generalization, we trained the same model architecture and compared its performance under \textbf{2} distinct conditions using the ID and OOD datasets as in Sec.~\ref{sec:data}. Following the data and training procedures described in Sec.~\ref{sec:datapreprocessing} and \ref{sec:training}, we had

\begin{itemize}
    \item \textbf{Baseline model}: The model was trained without any noise-based data augmentation
    \item \textbf{Noise-based Augmentation}: The model was trained with noise-based data augmentation (Gaussian, Speckle, Poisson, and Salt \& Pepper) applied randomly to each image in each epoch.
\end{itemize}
Both models were trained on the ID dataset. Subsequently, we evaluated their performance one last time on the ID and OOD test sets across 5 key metrics: \textbf{AUC}, \textbf{accuracy}, \textbf{F1}, \textbf{recall}, and \textbf{specificity}. 

\textbf{Ablation studies} While our primary experiment used BIMCV-COVID-19+ and Padchest as the ID dataset to simulate a single-source training scenario (Sec.~\ref{sec:dataselection}), we extended our analysis to examine scenarios where the limited ID data originated from diverse medical networks. We conducted three additional experiments by altering the composition of the ID sources to the following pairs: V2-COV19-NII and NIH, V2-COV19-NII and Padchest, and BIMCV-COVID-19+ and NIH. In these ablation studies, the OOD test sets comprised the remaining datasets not used in the respective ID source.

\section{RESULTS}

\begin{table}[ht]
\centering
\caption{Robustness Comparison: Absolute Generalization Gaps and Reduction in Gaps for Run 2 - the ID source comprises BIMCV-COVID-19+ and Padchest, while the OOD test set has COVID-19-AR, V2-COV19-NII, NIH, and Chexpert}
\label{tab:run2_robustness_results}
\begin{tabular}{lccc}
\toprule
\textbf{Metric} & \textbf{Gap (Without Noise)} & \textbf{Gap (With Noise)} & \textbf{Gap Reduction [95\% CI]} \\ 
\midrule
\textbf{AUROC}       & $0.060$ & $0.021$ & $\mathbf{0.039 \ [0.019, 0.059]^\dagger}$ \\ \addlinespace
\textbf{F1 Score}    & $0.032$ & $0.035$ & $-0.004 \ [-0.035, 0.028]$ \\ \addlinespace
\textbf{Accuracy}    & $0.059$ & $0.020$ & $\mathbf{0.039 \ [0.010, 0.068]^\dagger}$ \\ \addlinespace
\textbf{Recall}      & $0.097$ & $0.027$ & $\mathbf{0.069 \ [0.015, 0.124]^\dagger}$ \\ \addlinespace
\textbf{Specificity} & $0.046$ & $0.112$ & $-0.066 \ [-0.119, -0.014]$ \\ 
\bottomrule
\addlinespace[1ex]
\multicolumn{4}{l}{\small Gaps represent $|ID - OOD|$. Reduction is $Gap_{Without Noise} - Gap_{With Noise}$.} \\
\multicolumn{4}{l}{\small Positive values indicate gap reduction; negative values indicate gap increase.} \\
\multicolumn{4}{l}{\small $^\dagger$ denotes statistically significant gap reduction ($p < 0.05$).}
\end{tabular}
\end{table}

\begin{table}[ht]
\centering
\caption{Performance comparison on ID and OOD test sets between models trained with and without noise-based data augmentation across 5 key metrics: AUC, F1, accuracy, recall, and specificity. In this case, the ID source comprises BIMCV-COVID-19+ and Padchest, while the OOD test set has COVID-19-AR, V2-COV19-NII, NIH, and Chexpert. Metric values, rounded to 3 decimal places, are averaged over 31 runs with 31 different random seeds.}
\label{tab:performance_comparison_run2}
\begin{tabular}{llccc}
\toprule
\textbf{Metric} & \textbf{Approach} & \textbf{ID} & \textbf{OOD} & \textbf{Gap ($|\Delta_{ID-OOD}|$)} \\ 
\midrule
\textbf{AUROC}  & With Noise & $0.858 \pm 0.042$ & $0.879 \pm 0.033$ & $0.021 \ [0.002, 0.040]$ \\
                & Without Noise& $0.886 \pm 0.010$ & $0.827 \pm 0.018$ & $0.060 \ [0.052, 0.067]$ \\ 
\midrule
\textbf{F1 Score}  & With Noise & $0.807 \pm 0.060$ & $0.842 \pm 0.049$ & $0.035 \ [0.007, 0.063]$ \\
                & Without Noise& $0.805 \pm 0.021$ & $0.773 \pm 0.038$ & $0.032 \ [0.016, 0.047]$ \\ 
\midrule
\textbf{Accuracy}  & With Noise& $0.772 \pm 0.052$ & $0.792 \pm 0.050$ & $0.020 \ [-0.006, 0.046]$ \\
                & Without Noise& $0.773 \pm 0.019$ & $0.714 \pm 0.036$ & $0.059 \ [0.044, 0.074]$ \\ 
\midrule
\textbf{Recall}  & With Noise& $0.803 \pm 0.109$ & $0.775 \pm 0.082$ & $0.027 \ [-0.022, 0.076]$ \\
                & Without Noise& $0.770 \pm 0.042$ & $0.673 \pm 0.061$ & $0.097 \ [0.070, 0.123]$ \\ 
\midrule
\textbf{Specificity}  & With Noise& $0.725 \pm 0.098$ & $0.837 \pm 0.095$ & $0.112 \ [0.063, 0.161]$ \\
                & Without Noise& $0.778 \pm 0.042$ & $0.825 \pm 0.046$ & $0.046 \ [0.024, 0.068]$ \\ 
\bottomrule
\addlinespace[1ex]
\multicolumn{5}{l}{\small Values are reported as Mean $\pm$ SD. Gaps are reported as Absolute Mean $[95\% \text{ CI}]$.} \\ \\
\end{tabular}
\end{table}

For each experiment, the final metric values were averaged over 31 runs with 31 different random seeds.

When utilizing BIMCV-COVID-19+ and Padchest as primary ID sources, our empirical results demonstrate that noise injection, or noise-based data augmentation, could significantly improve model generalization to external data sources. As shown in Table~\ref{tab:run2_robustness_results}, our technique effectively halved the generalization gap for AUC, accuracy, and recall while recall remained relatively the same. 

Interestingly, in Table~\ref{tab:performance_comparison_run2}, F1 score was higher in OOD setting than ID by a margin of $0.037$ following noise augmentation. However, our ablation studies (Table~\ref{tab:performance_comparison}), in which we used a small consortium of unrelated medical networks for training, reveal a contrasting trend: Specificity exhibited a significant decline. We suspect that data composition might have played a key role in this phenomena as suggested by ref.~\citenum{ahmed2021discovery} that data from the same medical networks can mitigate the effects of shortcuts and, perhaps, facilitate learning reasonable biomarkers that can translate well across distributions.

Regarding our ablation studies, we altered the composition of our ID sources, where the datasets for both COVID-19 and Pneumonia originated from different medical networks: Run 1 -  V2-COV19-NII (Germany) and NIH (USA), Run 2 - V2-COV19-NII (Germany) and Padchest (Spain), and Run 3 - BIMCV-COVID-19+ (Spain) and NIH (USA). This approach helps us determine whether noise-based data augmentation is sensitive to data composition. Table~\ref{tab:full_robustness_results} reveals that the technique indeed improved model generalization compared to training without noise-based augmentation, but it is worth noting that there was still a dramatic gap between ID and OOD evaluation in either case. It suggests that, given a small dataset, the data composition itself plays a pivotal role in guiding the model to learn reasonable biomarkers that can be translated to a new distribution and adapt to domain gaps. Noise-based data augmentation can then significantly enhance the model's generalization to data from unseen sources and give the model end users a reasonable expectation of its performance under a distribution shift. When the dissimilarity between the ID sources begins to become sufficiently large, however, classic noise-based augmentation techniques may not suffice and models may continue relying on shortcuts to maximize their ID performance.

\section{CONCLUSIONS}
In conclusion, given a limited training dataset in size and source diversity, noise-based data augmentation can improve our DL model generalization to the same pathology from different sources not covered in the training set. On the other hand, data composition is worth thorough consideration as it plays a pivotal role in helping the model learn generalizable, reasonable biomarkers that remain valid under a distribution shift. 
\bibliography{report} % bibliography data in report.bib
\bibliographystyle{spiebib} % makes bibtex use spiebib.bst

\appendix

\begin{table}[ht]
\centering
\caption{Performance comparison on ID and OOD test sets between models trained with and without noise-based data augmentation across AUC, F1, accuracy, recall, and specificity. In this table, our ID sources alternate between Run 1 -  V2-COV19-NII (Germany) and NIH (USA), Run 3 - V2-COV19-NII (Germany) and Padchest (Spain), and Run 4 - BIMCV-COVID-19+ (Spain) and NIH (USA). The OOD test set are the remaining data sources not covered in each respective ID source. The final results are averaged over 31 runs with 31 different random seeds.}
\label{tab:performance_comparison}
\begin{tabular}{llccc}
\toprule
\textbf{Metric} & \textbf{Approach} & \textbf{ID} & \textbf{OOD} & \textbf{Gap ($|\Delta_{ID-OOD}|$)} \\ 
\midrule
\textbf{AUROC}  & With Noise Run 1 & $0.905 \pm 0.029$ & $0.774 \pm 0.039$ & $0.131 \ [0.114, 0.149]$ \\
                & Without Noise Run 1& $0.925 \pm 0.016$ & $0.747 \pm 0.052$ & $0.178 \ [0.158, 0.198]$ \\ \\
                
                & With Noise Run 3 & $0.971 \pm 0.017$ & $0.768 \pm 0.022$ & $0.203 \ [0.193, 0.213]$ \\
                & Without Noise Run 3& $0.999 \pm 0.002$ & $0.721 \pm 0.028$ & $0.278 \ [0.268, 0.288]$ \\ \\
                
                & With Noise Run 4 & $0.790 \pm 0.067$ & $0.713 \pm 0.080$ & $0.077 \ [0.040, 0.115]$ \\
                & Without Noise Run 4& $0.895 \pm 0.020$ & $0.661 \pm 0.059$ & $0.233 \ [0.210, 0.256]$ \\ 
\midrule
\textbf{F1 Score}  & With Noise Run 1 & $0.924 \pm 0.026$ & $0.832 \pm 0.004$ & $0.091 \ [0.082, 0.101]$ \\
                & Without Noise Run 1& $0.900 \pm 0.023$ & $0.834 \pm 0.004$ & $0.066 \ [0.057, 0.075]$ \\ \\
                
                & With Noise Run 3 & $0.949 \pm 0.010$ & $0.801 \pm 0.004$ & $0.148 \ [0.144, 0.152]$ \\
                & Without Noise Run 3& $0.960 \pm 0.015$ & $0.800 \pm 0.004$ & $0.160 \ [0.155, 0.166]$ \\ \\
                
                & With Noise Run 4 & $0.602 \pm 0.106$ & $0.792 \pm 0.098$ & $0.191 \ [0.139, 0.243]$ \\
                & Without Noise Run 4& $0.699 \pm 0.045$ & $0.779 \pm 0.056$ & $0.080 \ [0.054, 0.106]$ \\ 
\midrule
\textbf{Accuracy}  & With Noise Run 1 & $0.896 \pm 0.033$ & $0.723 \pm 0.010$ & $0.172 \ [0.160, 0.185]$ \\
                & Without Noise Run 1& $0.867 \pm 0.029$ & $0.726 \pm 0.009$ & $0.141 \ [0.130, 0.152]$ \\ \\
                
                & With Noise Run 3 & $0.928 \pm 0.015$ & $0.688 \pm 0.006$ & $0.240 \ [0.234, 0.246]$ \\
                & Without Noise Run 3& $0.944 \pm 0.022$ & $0.694 \pm 0.007$ & $0.250 \ [0.241, 0.258]$ \\ \\
                
                & With Noise Run 4 & $0.612 \pm 0.138$ & $0.710 \pm 0.083$ & $0.097 \ [0.039, 0.156]$ \\
                & Without Noise Run 4& $0.732 \pm 0.058$ & $0.685 \pm 0.056$ & $0.048 \ [0.019, 0.077]$ \\ 
\midrule
\textbf{Recall}  & With Noise Run 1 & $0.952 \pm 0.049$ & $0.986 \pm 0.007$ & $0.034 \ [0.016, 0.052]$ \\
                & Without Noise Run 1& $0.905 \pm 0.042$ & $0.989 \pm 0.007$ & $0.084 \ [0.068, 0.099]$ \\ \\
                
                & With Noise Run 3 & $0.998 \pm 0.009$ & $0.920 \pm 0.021$ & $0.079 \ [0.070, 0.087]$ \\
                & Without Noise Run 3& $1.000 \pm 0.000$ & $0.892 \pm 0.019$ & $0.108 \ [0.101, 0.115]$ \\ \\
                
                & With Noise Run 4 & $0.842 \pm 0.188$ & $0.800 \pm 0.179$ & $0.042 \ [-0.051, 0.136]$ \\
                & Without Noise Run 4& $0.890 \pm 0.054$ & $0.765 \pm 0.100$ & $0.125 \ [0.084, 0.166]$ \\ 
\midrule
\textbf{Specificity}  & With Noise Run 1 & $0.784 \pm 0.045$ & $0.121 \pm 0.047$ & $0.663 \ [0.640, 0.687]$ \\
                & Without Noise Run 1& $0.790 \pm 0.030$ & $0.120 \pm 0.036$ & $0.670 \ [0.653, 0.687]$ \\ \\
                
                & With Noise Run 3 & $0.787 \pm 0.043$ & $0.188 \pm 0.054$ & $0.599 \ [0.574, 0.623]$ \\
                & Without Noise Run 3& $0.832 \pm 0.065$ & $0.268 \pm 0.056$ & $0.564 \ [0.533, 0.595]$ \\ \\
                
                & With Noise Run 4 & $0.492 \pm 0.261$ & $0.452 \pm 0.263$ & $0.039 \ [-0.094, 0.172]$ \\
                & Without Noise Run 4& $0.649 \pm 0.100$ & $0.455 \pm 0.096$ & $0.194 \ [0.144, 0.244]$ \\ 
\bottomrule
\addlinespace[1ex]
\multicolumn{5}{l}{\small Values are reported as Mean $\pm$ SD. Gaps are reported as Absolute Mean $[95\% \text{ CI}]$.} \\
\end{tabular}
\end{table}

\begin{table}[ht]
\centering
\caption{Robustness Comparison: Absolute Generalization Gaps and Reduction in Gaps Across Experimental Runs}
\label{tab:full_robustness_results}
\begin{tabular}{lcccc}
\toprule
\textbf{Metric} & \textbf{Experiment} & \textbf{Gap (Without Noise)} & \textbf{Gap (With Noise)} & \textbf{Gap Reduction [95\% CI]} \\ 
\midrule
\textbf{AUROC}       & Run 1 & $0.178$ & $0.131$ & $\mathbf{0.047 \ [0.021, 0.072]^\dagger}$ \\
                     & Run 3 & $0.278$ & $0.203$ & $\mathbf{0.075 \ [0.062, 0.089]^\dagger}$ \\ 
                     & Run 4 & $0.233$ & $0.077$ & $\mathbf{0.156 \ [0.113, 0.199]^\dagger}$ \\
\midrule
\textbf{F1 Score}    & Run 1 & $0.066$ & $0.091$ & $-0.025 \ [-0.037, -0.013]$ \\
                     & Run 3 & $0.160$ & $0.148$ & $\mathbf{0.013 \ [0.006, 0.020]^\dagger}$ \\ 
                     & Run 4 & $0.080$ & $0.191$ & $-0.111 \ [-0.168, -0.054]$ \\
\midrule
\textbf{Accuracy}    & Run 1 & $0.141$ & $0.172$ & $-0.031 \ [-0.047, -0.015]$ \\
                     & Run 3 & $0.250$ & $0.240$ & $0.010 \ [-0.000, 0.020]$ \\ 
                     & Run 4 & $0.048$ & $0.097$ & $-0.050 \ [-0.113, 0.014]$ \\
\midrule
\textbf{Recall}      & Run 1 & $0.084$ & $0.034$ & $\mathbf{0.050 \ [0.027, 0.073]^\dagger}$ \\
                     & Run 3 & $0.108$ & $0.079$ & $\mathbf{0.029 \ [0.019, 0.040]^\dagger}$ \\ 
                     & Run 4 & $0.125$ & $0.042$ & $0.083 \ [-0.017, 0.183]$ \\
\midrule
\textbf{Specificity} & Run 1 & $0.670$ & $0.663$ & $0.007 \ [-0.022, 0.035]$ \\
                     & Run 3 & $0.564$ & $0.599$ & $-0.035 \ [-0.073, 0.004]$ \\ 
                     & Run 4 & $0.194$ & $0.039$ & $\mathbf{0.155 \ [0.015, 0.294]^\dagger}$ \\
\bottomrule
\addlinespace[1ex]
\multicolumn{5}{l}{\small Gaps represent $|ID - OOD|$. Reduction is $|Gap_{Without Noise}| - |Gap_{With Noise}|$.} \\
\multicolumn{5}{l}{\small $^\dagger$ denotes statistically significant gap reduction ($p < 0.05$)}
\end{tabular}
\end{table}

\end{document}